\documentclass[12pt, a4paper]{article}

\usepackage[utf8]{inputenc}
\usepackage[T1]{fontenc}
\usepackage{mathptmx}
\usepackage[margin=1in]{geometry}
\usepackage{setspace}
\usepackage{titlesec}
\usepackage{fancyhdr}
\usepackage{hyperref}
\usepackage{csquotes}
\usepackage{microtype}
\usepackage{hanging}
\usepackage{abstract}
\usepackage{booktabs}
\usepackage{array}
\usepackage{parskip}
\usepackage{natbib}
\usepackage{amsmath}
\usepackage{enumitem}
\usepackage{xcolor}
\usepackage{soul}

\fancypagestyle{plain}{\fancyhf{}}

\titleformat{\section}{\normalfont\bfseries}{\thesection.}{0.5em}{}
\titleformat{\subsection}{\normalfont\bfseries\itshape}{\thesubsection}{0.5em}{}
\titlespacing*{\section}{0pt}{2ex}{0.5ex}
\titlespacing*{\subsection}{0pt}{1.5ex}{0.3ex}

\hypersetup{colorlinks=true, linkcolor=black, citecolor=black, urlcolor=black}

\begin{document}

\begin{titlepage}
  \centering
  \vspace*{1.5in}
  {\large\bfseries Artificial Intelligence as Monism: \\[0.4em]
  Ontological, Organisational, and Methodological Implications\par}
  \vspace{1.5em}
  {Bertrand K. Hassani\par}
  \vspace{0.5em}
 \footnotesize{QUANT AI Lab, C. de Arturo Soria, 122, 28043 Madrid, Spain, Department of Computer Science, University College London, Gower Street, London WC1E 6EA, UK, Centre d'Économie de la Sorbonne (CES), Maison des Sciences Économiques (MSE), Université Paris 1 Panthéon-Sorbonne, 106-112 Boulevard de l'Hôpital, 75013 Paris, France, Institut Louis Bachelier, Palais Brongniart, 28 Place de la Bourse, 75002 Paris, France\par}
  \vspace{0.5em}
  {June 2026\par}
  \vspace{2em}
  \begin{abstract}
  \noindent
  This paper argues that Artificial Intelligence should be understood as a form of monism:
  a unified substance that cannot be decomposed into separate elements such as data,
  algorithms, or technical architectures. Drawing from philosophical traditions of monism,
  dualism, and holism, the paper contends that AI is not merely a collection of components
  but a single, indivisible essence reflecting the phenomena it replicates. Treating AI as
  monism has deep implications across multiple dimensions. Epistemologically, it positions
  AI as the central interpretive force across technological, organisational, and societal
  domains, while raising ethical and existential concerns regarding singularity, the
  homogenisation of innovation, and the concentration of decision-making power. At the
  organisational level, a monistic approach challenges traditional siloed structures,
  advocating instead for transversal, problem-centric teams whose mandate derives from
  the integrity of the problem rather than from departmental hierarchy. In project
  management, it implies a unified vision and an integrated evaluation of complexity in
  which no single stakeholder perspective dominates the assessment of outcomes. In data
  and information management, it calls for architectures that reflect the irreducible unity
  of the phenomena being modelled. Ultimately, this paper calls for a paradigm shift in
  how AI is conceptualised, governed, and integrated, suggesting that only by embracing AI
  as monism can organisations achieve genuine agility and avoid the structural
  inefficiencies inherent to reductionist approaches.

  \vspace{1em}
  \end{abstract}
\end{titlepage}

\newpage

\section{Introduction}

\noindent What is Artificial Intelligence? It is a question deceptive in its apparent simplicity, and yet the answers it elicits reveal far more about the respondent than about the subject itself. To ask the infrastructure architect is to receive an answer framed in systems,
pipelines, and cloud deployments; to ask the data scientist is to hear of models,
training sets, and loss functions; to ask the person in the street is to be told, with
increasing confidence, that it is ChatGPT. Each answer is partial, locally coherent, and
systemically misleading. This paper argues that the persistent inability to arrive at a
satisfactory, unified definition of AI is not an accident of nomenclature or the product
of disciplinary fragmentation alone. It reflects a deeper conceptual error: the assumption
that AI can be meaningfully decomposed into constituent parts at all.\\

\noindent The standard definition treats AI as the simulation of human intelligence in machines, specifically the capacities to learn, reason, perceive, and interact with an
environment~\citep{mccarthy2006proposal}. This definition is simultaneously too narrow
and too broad. It is too narrow in that it reduces human intelligence to a set of
functional capacities that can be discretely enumerated; it is too broad in that, if AI
encompasses the full range of these capacities, it encompasses essentially everything.
The definition satisfies neither philosophical rigour nor practical utility.\\

\noindent The position advanced in this paper is that AI should be understood as a \textit{monism}~\citep{schaffer2010monism}: a single, unified substance or principle that cannot be meaningfully separated into component parts such as data, algorithms, or technical architectures, because those parts are not independent entities in the first place. They are aspects of a single underlying reality, namely the phenomenon that AI is attempting to model. A financial default, a fraudulent transaction, a purchasing decision, these are not composites of data plus algorithms plus infrastructure; they are indivisible events in the world. Any adequate representation of them must preserve that indivisibility rather than fracture it across departmental silos and technical layers.\\

\noindent The paper proceeds as follows. Section~2 provides a philosophical grounding of the
concept of monism and distinguishes it from the related but distinct notions of dualism
and holism. Section~3 examines the broader implications of treating AI as a monistic
force, including its ethical and existential dimensions. Section~4 develops the technical
and methodological case for AI as monism and draws out its consequences for organisational
design. Section~5 addresses the implications for project management. Section~6 considers
the consequences for data governance and information architecture. Section~7 concludes.\\

\section{Philosophical Foundations: Monism, Dualism, and Holism}

\noindent Before the concept of AI as monism can be examined in its organisational and technical dimensions, it is necessary to be precise about what monism means and how it differs from the adjacent philosophical positions of dualism and holism. These distinctions are not merely terminological: they carry fundamentally different consequences for how AI systems should be designed, governed, and institutionally embedded.\\

\subsection{Monism}

\noindent Ontological monism is the philosophical position that reality is fundamentally composed of a single substance, principle, or kind of thing. In contrast to dualism, which posits two fundamental kinds of substance, or to pluralism, which posits
many, monism asserts that everything that exists is ultimately reducible to or rather
expressive of one underlying reality or essence. The history of Western philosophy
offers multiple variants: Spinoza's substance monism, in which God or Nature constitutes
the single infinite substance of which mind and matter are merely attributes; neutral
monism, associated with Bertrand Russell and William James, which posits a single neutral
thing that is neither mental nor physical; and physicalist monism, which holds that
everything is ultimately physical~\citep{spinoza2024ethics, james1904does}.\\

\noindent The relevance of monism to AI is not primarily metaphysical in the traditional sense. Rather, it is applied at the level of what might be called \textit{representational
ontology}: the claim that AI systems, the data they consume, the algorithms they embody,
and the architectures on which they run are not three separate entities that happen to be
combined, but expressions of a single underlying phenomenon, i.e. the real-world event or
process being modelled. In this sense, the analogy is apt: just as, in a monistic
ontology, whatever exists is composed of one substance regardless of how it is divided or
described, so an AI system and the phenomenon it models are, at the level of essence,
the same thing. To separate them, as conventional practice does, by maintaining
independent teams for data, for algorithms, and for infrastructure, is to commit a
category error.\\

\subsection{Dualism and its Limitations}

\noindent Dualism, most famously articulated by Descartes as the separation of \textit{res cogitans} (the thinking substance) and \textit{res extensa} (the extended, material substance), maintains that reality consists of two irreducibly distinct kinds of thing~\citep{descartes2016meditations}. The conventional treatment of AI implicitly operates within a dualist framework: there is the "technical" side of AI (data, algorithms, infrastructure) and the "business" side (the problem to be solved, the organisational objective, the regulatory context). These two sides are understood to interact but are managed, budgeted, staffed, and evaluated independently.\\

\noindent The dualist framing produces the inefficiencies that practitioners routinely observe: data teams that deliver outputs that algorithms cannot consume; algorithmic solutions that cannot be deployed because the infrastructure is not ready; infrastructure that is technically sound but architecturally misaligned with the problem it is meant to support. Each of these failures is a consequence of treating as separate what is in fact one.\\

\subsection{Holism and its Distinction from Monism}

\noindent Holism is the view that a system or entity possesses properties that cannot be
explained by, or reduced to, the properties of its parts alone that the whole is
greater than the sum of its parts~\citep{smuts1926holism}. Holism has become a
fashionable orientation in discussions of AI governance and organisational design, and it
is a genuine improvement over the fragmented, reductionist approaches that have dominated
practice. However, holism is not monism, and the distinction matters.\\

\noindent A holist view of AI would maintain that while data, algorithms, and infrastructure are distinct entities, they must be considered together in order to understand the behaviour of the whole. The parts retain their distinct identities; what changes is the analytical frame. A monist view goes further: the parts do not merely need to be considered together; at the level of essence, they are not distinct entities at all. The fraud that an AI model seeks to detect is not constituted by a dataset, an algorithm, and a cloud deployment that happen to be combined; it is a single event in the world, and the model that represents it is, in essence, that event. The practical consequence is that
the architectural and institutional separations that a holist approach would still
permit maintaining distinct teams, distinct budgets, distinct evaluation frameworks,
while encouraging their coordination are precisely what a monist approach disallows.\\

\section{Implications of AI as Monism}

\noindent The treatment of AI as a monistic force carries implications that range from the
organisational and methodological to the ethical and existential. This section addresses
each in turn.\\

\subsection{Technological Integration and the Risk of Monoculture}

\noindent If AI is understood as the unified force driving innovation across all domains, from medicine and finance to education and public administration, then the natural
consequence is that it becomes the governing principle of technological development. This
has an obvious positive dimension: the convergence of formerly disparate fields around a
common framework accelerates the transfer of insights and the accumulation of
capabilities. The application of deep learning architectures developed for natural
language processing to protein structure prediction~\citep{jumper2021highly} is one
illustration of how a monistic orientation can generate unexpected and transformative
cross-domain gains.\\

\noindent The negative dimension, however, deserves equal attention. If AI is the singular lens through which all problems are filtered, the diversity of approaches that has historically been a source of scientific and technological resilience is at risk. A monoculture of method is as fragile as a monoculture of crop: optimal under the conditions for which it was cultivated, and catastrophically vulnerable to conditions it was not designed to handle. The homogenisation of innovation, the tendency, already observable, for AI systems to converge on similar architectures, similar training regimes, and similar
evaluation criteria may produce short-term efficiency gains at the cost of long-term
adaptive capacity~\citep{bender2021dangers}.\\

\subsection{Power, Accountability, and the Concentration of Decision-Making}

\noindent If AI becomes the singular lens through which information and decisions are processed and filtered, it necessarily centralises power. The question of who designs, trains, and controls the AI becomes, under a monistic framing, the question of who shapes reality itself. This is not a metaphorical concern; it is an operational one. The biases embedded in training data, the objective functions encoded in models, and the deployment decisions made by a small number of organisations with disproportionate access to computational resources and proprietary data are, under a monistic treatment of AI, constitutive of the reality that AI models and mediates.\\

\noindent Accountability in this context becomes genuinely difficult. In a dualist or pluralist framework, it is at least conceptually possible to assign responsibility to specific components: the dataset that introduced a demographic bias, the algorithm that
amplified it, the deployment decision that made it consequential. In a monist framework,
where the model is the phenomenon it models, the attribution of causal responsibility to
any single component is a category error. This presents both a theoretical challenge for
legal and regulatory frameworks and a practical challenge for governance
design~\citep{mittelstadt2016ethics}.\\

\subsection{The Singularity Hypothesis and Existential Risk}

\noindent The singularity hypothesis, i.e., the proposition that AI will at some point reach a threshold of self-improvement beyond which its growth becomes self-perpetuating and no longer subject to human direction is, under a monistic reading of AI, not merely a
speculative technological scenario but the logical endpoint of treating AI as a unified,
self-contained substance~\citep{vinge1993coming, kurzweil2005singularity}. If AI is, in
essence, the phenomenon it models, and if that phenomenon is human intelligence itself,
then the development of AI is, in monistic terms, the development of intelligence as
such, with no principled distinction between artificial and natural instances.\\

\noindent The existential concern is not, however, the singularity \textit{per se}, but rather the \textit{premature assumption} of singularity: the practical and institutional treatment of AI as if it had already achieved the comprehensive, self-sufficient intelligence that singularity implies, leading to the uncritical delegation of consequential decisions to systems that remain, in reality, bounded, brittle, and deeply dependent on the assumptions embedded in their training~\citep{marcus2019rebooting}. The monist framing, if adopted carelessly, risks providing an intellectual justification for precisely this premature delegation.\\

\subsection{Human Autonomy and the Standardisation of Judgment}

\noindent The most immediately consequential implication of AI as a monistic force is its effect on human judgment and autonomy. If AI is treated as the unified, sovereign method for
processing information and making decisions, the diversity of human judgment, with its
contextual sensitivity, its affective dimensions, its moral weight, is not supplemented
but displaced~\citep{floridi2018ai4people}. The concern is not the familiar one about
job replacement; it is more fundamental. It is the concern that the variety of ways in
which human beings understand and navigate the world, the plurality of epistemic and
moral frameworks that different cultures, disciplines, and individuals bring to bear on
shared problems, may be progressively narrowed by the convergence of all information
processing on a single, algorithmically determined framework.\\

\noindent This concern is directly relevant to the ethical dimensions of AI governance. Moral diversity, i.e., the existence of multiple, sometimes incompatible, frameworks for evaluating human action, is not a problem to be solved by AI; it is a feature of the moral
landscape that any adequate governance framework must accommodate. A monistic AI that
attempts to enforce a universal ethical standard will clash with the legitimate plurality
of moral traditions; one that attempts to accommodate all moral systems simultaneously
will face contradictions that cannot be resolved algorithmically~\citep{russell2022human}.\\

\section{Technical and Methodological Monism: Organisational Implications}

\subsection{The Unity of Data, Algorithms, and Architecture}

\noindent The preceding sections have established the philosophical and ethical dimensions of AI as monism. The present section descends to the level of practical consequence. The central claim is that the conventional separation of AI projects into distinct streams, i.e. data management, algorithm development, and technical architecture, is not merely
organisationally inconvenient but epistemologically incoherent.\\

\noindent Consider the canonical example of credit default modelling. A model of corporate default is understood, in standard practice, to consist of a dataset (financial statements,
market prices, macro-economic indicators), an algorithm (a logistic regression, a
gradient-boosted tree, a neural network), and an infrastructure (a database, a compute
cluster, a deployment pipeline). Each of these components is managed by a different team,
evaluated by different criteria, and governed by different institutional processes.\\

\noindent The monistic view holds that this decomposition is a representation error. A corporate default is not a combination of a dataset, an algorithm, and an infrastructure; it is a singular event in the world, embedded in a network of economic relationships, regulatory expectations, managerial decisions, and market dynamics that constitute it as a whole. An AI model that adequately represents a corporate default must preserve that wholeness, not decompose it into tractable but ultimately artificial sub-problems. When the dataset team optimises for completeness and coverage, the algorithm team optimises for predictive accuracy on a held-out test set, and the infrastructure team optimises for deployment latency, they are each optimising for a partial representation of the phenomenon, and the result of combining these partial representations is not the whole but a composite of misalignments~\citep{sculley2015hidden}.\\

\noindent The same observation applies, with equal force, to fraud detection, operational risk modelling, customer behaviour prediction, and any other domain in which AI is applied to real-world phenomena. The phenomenon is one; the model should be one; and the process
by which the model is developed, validated, and deployed should reflect that unity.\\

\subsection{The Incompatibility of Silos with Monistic AI}

\noindent The conventional organisational architecture of technology-intensive institutions is hierarchical and siloed. Under a typical corporate structure, a Chief Executive Officer
oversees a set of functional divisions, for instance, Finance, Marketing, Risk, Human Resources, Technology, Data, and increasingly, AI, each of which maintains its own budget,
objectives, performance metrics, and reporting lines. This architecture is not accidental;
it reflects a dualist or pluralist ontology in which the business problem, the data, the
algorithm, and the infrastructure are genuinely distinct entities that happen to need
coordination.\\

\noindent If AI is a monism, then this architecture is not merely suboptimal but
structurally incompatible with the nature of the subject matter. Silos are mechanically incompatible with monism, because monism admits of no principled partition. The distinctions that silos enforce between the data team and the algorithm team, between the technology function and the risk function, between the model developer and the model deployer, are, from a monistic perspective, arbitrary divisions of an indivisible whole, and as such they systematically introduce distortions into the representation of the phenomena being modelled.\\

\noindent The organisational consequence of taking AI as monism seriously is therefore radical: the silos must be dissolved, not merely connected or coordinated. The unit of organisational analysis and the unit of resource allocation should be not the function but the problem. Each problem to be addressed, each phenomenon to be modelled , warrants a dedicated, cross-functional team whose mandate is defined by the integrity of the problem rather than by departmental affiliation. In such a structure, the project manager acquires a
genuinely transversal authority: not the authority of a coordinator who negotiates between
functional representatives, but the authority of a sovereign who is accountable for the
quality of the solution as a whole.\\

\noindent This is not a utopian or impractical proposal. It is, in fact, closer to the operating model of high-performing technology organisations such as Amazon's two-pizza team
model~\citep{bryar2021working}, the squad structure adopted by spotify~\citep{kniberg2012scaling} or the product-centric operating model increasingly adopted by leading financial institutions, than to the functional hierarchy that most traditional organisations still maintain. The argument from AI as monism provides a principled justification for these organisational forms, and suggests that the observed difficulty of industrialising AI solutions in large institutions is not primarily a technical problem but a structural one: the architecture of the organisation is misaligned with the nature of what it is trying to build.\\

\subsection{Towards a Transversal Organisational Architecture}

\noindent The organisational architecture implied by AI as monism can be characterised along four dimensions. First, it is \textit{problem-centric}: teams are constituted around problems, not functions, and are dissolved when the problem is resolved or when it evolves to the point that a different team composition is required. Second, it is \textit{transversal}: all skills necessary to address the problem in its unity, i.e., domain expertise, data engineering, algorithm development, infrastructure, legal and compliance, and communication, are present within the team from the outset, not added incrementally as handoff stages. Third, it is \textit{outcome-accountable}: the team is evaluated on the quality of the solution in relation to the real-world phenomenon being modelled, not on the internal quality of its component outputs. Fourth, it is \textit{authority-unified}: the project manager holds genuine decision authority over all dimensions of the project, not merely over the coordination of functional contributions.\\

\noindent Such an architecture is not without its own risks. The dissolution of functional
hierarchies removes the institutional memory and accumulated expertise that those
hierarchies, at their best, preserve. The concentration of authority in the project
manager introduces new accountability questions and may create new forms of bottleneck.
These are genuine concerns that a mature implementation of monistic organisational design
must address. The claim is not that the transversal, problem-centric model is without
cost, but that the costs of maintaining the siloed, function-centric model in the context
of AI development are substantially higher and less visible.\\

\section{Project Management Under a Monistic Framework}

\subsection{The Centralisation of Vision}

\noindent If a project is managed under a monist view, the centralisation of vision is a mechanical consequence. One governing principle, defined not by a single stakeholder's preference but by the integrity of the phenomenon being modelled, dominates. All decisions align to that single axis. This does not mean that complexity is ignored or that the interests of different stakeholders are disregarded; it means that complexity and diverse interests are assessed from the perspective of the whole rather than from the perspective of any individual component.\\

\noindent The contrast with conventional project management is instructive. Standard project
management frameworks - the Project Management Body of Knowledge~\citep{project2020guide}, and its successors - organise projects around a set of distinct knowledge
areas: scope, time, cost, quality, risk, resources, communications, procurement, and
stakeholder management. Each of these is treated as a separable domain requiring its own
planning, monitoring, and control processes. The result is a mode of project governance
that is analytically tractable but inherently fragmented: the risk register does not talk
to the resource plan; the quality assurance process does not inform the cost baseline;
the communications plan does not shape the scope definition.\\

\noindent A monistic project management framework treats these not as distinct domains to be
managed in parallel but as aspects of a single, indivisible challenge. Scope, time, cost,
and quality are not independent variables; they are interdependent dimensions of the
phenomenon being addressed, and changes to any one of them constitute changes to the
whole. The appropriate management response to a scope change is, therefore, not to assess
its impact on time, cost, and quality separately and then aggregate those assessments,
but to re-evaluate the project as a whole in light of the change and determine whether
the original governing principle is still being served.\\

\subsection{Evaluation and Key Performance Indicators}

\noindent The evaluation of project success under a monistic framework is necessarily different from conventional practice. Standard evaluation frameworks treat project success as a composite of sub-domain performance: a project is successful if it is delivered on time, within budget, to the specified quality, with acceptable risk outcomes. Each of these
criteria is assessed independently, and the project is judged successful if it meets a
sufficient number of them.\\

\noindent A monistic framework holds that this composite evaluation is incoherent, because the sub-domain criteria are not independent. A project delivered on time and within budget
but at a quality that fails to adequately represent the phenomenon being modelled is not
a successful project, regardless of its performance on the temporal and financial
dimensions. The only meaningful criterion of project success, under a monistic view, is
whether the delivered solution adequately represents the real-world phenomenon it was
designed to model, and whether that representation is robust enough to be practically
useful in the conditions under which it will be deployed.\\

\noindent Long-term outcome quality should therefore be the sole class of key performance
indicator that matters. Short-term surrogates, such as adherence to timeline, budget
compliance, test coverage, are useful diagnostics but must not be allowed to substitute
for the primary criterion. This has direct consequences for project governance: it
requires that evaluation processes extend beyond the project delivery date, that the
teams responsible for developing solutions retain accountability for their real-world
performance, and that the institutional incentives that currently reward on-time,
on-budget delivery regardless of outcome quality are systematically revised.\\

\subsection{Creativity, Arbitrage, and the Role of the Project Manager}

\noindent One concern about the monistic project management framework is that the centralisation of vision might stifle creativity by privileging the governing principle at the expense of alternative perspectives. This concern, while understandable, reflects a
misunderstanding of what monism requires. A monistic framework does not exclude
alternative viewpoints; it integrates them. If technical, human, social, and strategic
considerations have not been integrated into the governing vision, then the project is
not, in fact, being managed as a monism: the governing principle has not been
adequately defined.\\

\noindent The role of the project manager, in this context, is not to suppress alternative
perspectives but to perform the arbitrage function that resolves conflicts between them
in a manner consistent with the integrity of the whole. This requires a qualitatively
different kind of project management capability from that demanded by conventional
frameworks: not merely the ability to coordinate and track, but the ability to hold the
totality of the problem in view and to make decisions that are accountable to that
totality rather than to any single stakeholder's preference. The project manager, under
a monistic framework, is not a coordinator but a sovereign of the problem space.\\

\section{Data Governance and Information Architecture Under Monism}

\subsection{The Inadequacy of Conventional Data Architectures}

\noindent The conventional architecture of data management in large organisations reflects the same dualist and pluralist assumptions that characterise conventional organisational design. Data is treated as a resource that is distinct from the processes that consume it and from the systems that produce it. Data lakes, data warehouses, and data pipelines are
designed to accumulate and transform data independently of the specific phenomena being
modelled, on the assumption that a sufficiently comprehensive and well-governed data
resource will support any analytical requirement.\\

\noindent This assumption is inconsistent with a monistic understanding of AI. If data, algorithms, and the phenomena they model are aspects of a single underlying reality, then data cannot be meaningfully separated from the context of its use. A dataset of financial transactions is not raw material that can be fed into any algorithm for any purpose; it is an expression of a particular set of economic relationships, regulatory conditions, and
institutional behaviours, and its adequacy as a representation of those relationships
depends on whether it is used in a manner consistent with the phenomenon it encodes.\\

\subsection{Data Mesh and the Direction of Travel}

\noindent Recent developments in data architecture, in particular the data mesh paradigm
introduced by Dehghani~\citep{dehghani2022data}, represent a move in the direction
that a monistic framework implies, though they do not go far enough. Data mesh proposes
a decentralised approach in which data ownership is shifted from a central data team to
the domain teams that are closest to the phenomena being modelled. Each domain team is
responsible for the quality, accessibility, and governance of the data it produces and
consumes, supported by a domain-agnostic data platform team.\\

\noindent This is a significant improvement over the centralised data lake model, because it aligns data governance with the substantive knowledge of the domain teams and reduces the
systematic distortions that arise when data is managed by a team that is institutionally
separated from the phenomena it encodes. However, data mesh remains, in its canonical
formulation, a pluralist architecture: it posits multiple, domain-specific data products
that are related by common standards and a shared infrastructure, but it does not question
the fundamental separability of data from the phenomena it represents.\\

\noindent A monistic data architecture would go further. It would require that the design of data systems be driven not by considerations of data quality or data completeness in the
abstract, but by the specific requirements of the phenomena being modelled. It would
require that the teams responsible for data governance include, from the outset, the
domain expertise necessary to assess whether a given dataset adequately represents the
phenomenon it encodes. And it would require that the evaluation of data quality be
conducted not against abstract standards of completeness and consistency, but against the
criterion of representational adequacy: does this data, used in this way, enable a model
that genuinely captures the phenomenon in its unity?\\

\subsection{The Unity of Representation and the Problem of Decomposition}

\noindent The deepest implication of monistic data governance is what might be called the
\textit{problem of decomposition}: the tendency, deeply embedded in both the practice
and the theory of data management, to decompose complex real-world phenomena into
manageable sub-components that can be measured, stored, and processed independently.
This decomposition is epistemologically necessary for tractability, but it introduces
systematic distortions when the sub-components are treated as if they were genuinely
independent entities rather than aspects of an indivisible whole.\\

\noindent A financial default is not a combination of a debt-to-equity ratio, a credit rating, and a market price; it is a singular event constituted by a specific configuration of economic, managerial, and market conditions that, when decomposed into independent variables, loses precisely the relational properties that make it a default rather than a collection of financial indicators. The model that best represents a financial default is not the one that most accurately predicts each indicator in isolation, but the one that most
adequately captures the relational structure of the phenomenon as a whole.\\

\noindent This is not an argument against quantification or formalisation. It is an argument for a mode of quantification and formalisation that is attentive to the relational structure of the phenomena being modelled, and that does not mistake the convenience of
decomposition for the truth of independence. Building AI systems that honour this
constraint is technically demanding and institutionally challenging. But it is, the
argument of this paper suggests, the only mode of AI development that is genuinely
adequate to the nature of the subject matter.\\

\section{Conclusion}

\noindent This paper has argued that Artificial Intelligence should be understood as a monism: a unified substance that cannot be meaningfully decomposed into independent components such as data, algorithms, or technical architectures, because those components are not independent entities but aspects of a single underlying reality --- the real-world
phenomenon that AI is designed to model. This position has been developed through three
main lines of argument.\\

\noindent The first is philosophical. Drawing on the traditions of ontological monism, and
distinguishing the monistic position from the related but distinct positions of dualism
and holism, the paper has argued that the conventional treatment of AI as a set of
separable technical components reflects a dualist or pluralist ontology that is
inconsistent with the nature of the phenomena AI seeks to model. Real-world events ---
defaults, frauds, purchases, failures, are not composites of data, algorithms, and
infrastructure; they are singular occurrences, and models that decompose them into
independent components introduce systematic distortions from the outset.\\

\noindent The second is organisational. The siloed, function-centric organisational architectures that predominate in technology-intensive institutions are mechanically incompatible with the monistic understanding of AI, because they partition the development process along lines that do not correspond to any genuine distinction in the subject matter. The paper has proposed an alternative: a transversal, problem-centric organisational architecture in which teams are constituted around phenomena rather than functions, and in which the project manager holds genuine, unified authority over the solution as a whole.\\

\noindent The third is methodological. In both project management and data governance, the paper has argued that the monistic framework implies a fundamental reorientation of evaluation criteria: away from the compliance with sub-domain standards and toward the adequacy of the representation of the phenomenon as a whole. This reorientation is demanding, both technically and institutionally, but it is the only approach that is genuinely aligned with what AI development requires.\\

\noindent The paradigm shift called for by this paper will not be easy to implement. The
institutional, cultural, and commercial forces that sustain the siloed, reductionist
approach to AI development are powerful, and the short-term costs of the transition are
real. But the costs of not making the transition are higher, and less visible: the
systematic inability to industrialise AI solutions at scale, the persistent gap between
AI's demonstrated capability in controlled research environments and its performance in
production, and the deepening misalignment between the governance structures of
technology institutions and the nature of the technology they are trying to govern. AI
as monism is not a convenient metaphor. It is, this paper has argued, a necessary
reorientation.

\newpage
\bibliographystyle{apalike}
\bibliography{bibliography.bib}
\end{document}